%% file: main.tex
\documentclass{article}

\usepackage{microtype}
\usepackage{graphicx}
\usepackage{caption}
\usepackage{subcaption}
\usepackage{tikz}
\usepackage{amsmath}
\usepackage{amssymb}
\usepackage{amsthm}
\usepackage{dsfont}
\usepackage{tkz-graph}
\usepackage{amsthm}
\usepackage{tabularx}

\newcommand{\Real}{\mathbb{R}}

\usepackage{booktabs} % for professional tables

\usepackage{hyperref}

\usepackage[accepted]{icml2020}

\icmltitlerunning{Learning  Algebraic Multigrid  Using Graph Neural Networks}

\begin{document}

\twocolumn[
\icmltitle{Learning  Algebraic Multigrid  Using Graph Neural Networks}



\icmlsetsymbol{equal}{*}

\begin{icmlauthorlist}
\icmlauthor{Ilay Luz}{weiz}
\icmlauthor{Meirav Galun}{weiz}
\icmlauthor{Haggai Maron}{Nvidia}
\icmlauthor{Ronen Basri}{weiz}
\icmlauthor{Irad Yavneh}{tech}
\end{icmlauthorlist}

\icmlaffiliation{weiz}{Weizmann Institute of Science, Rehovot, Israel.}
\icmlaffiliation{tech}{Technion, Israel Institute of Technology, Haifa, Israel}
\icmlaffiliation{Nvidia}{NVIDIA Research}

\icmlcorrespondingauthor{Ilay Luz}{ilayluz@gmail.com}

\icmlkeywords{Machine Learning, ICML}

\vskip 0.3in
]


\printAffiliationsAndNotice{}  % leave blank if no need to mention equal contribution

\begin{abstract}
\label{abstract}
\input{abstract}
\end{abstract}

\section{Introduction}
\label{Introduction}
\input{introduction}

\section{Related work}
\label{relatedWork}
\input{relatedWork}

\section{AMG background}
\label{problemDefinition}
\input{problemDefinition}

\section{Learning Method}
\label{Method}
\input{method}

\section{Experiments}
\label{Experiments}
\input{experiments}

\bibliography{main}
\bibliographystyle{icml2020}

\end{document}

%% file: abstract.tex
Efficient numerical solvers for sparse linear systems are crucial in science and engineering. One of the fastest methods for solving large-scale sparse linear systems is algebraic multigrid (AMG). The main challenge in the construction of AMG algorithms is the selection of the prolongation operator---a problem-dependent sparse matrix which governs the multiscale hierarchy of the solver and is critical to its efficiency. Over many years, numerous methods have been developed for this task, and yet there is no known single right answer except in very special cases. Here we propose a framework for learning AMG prolongation operators for linear systems with sparse symmetric positive (semi-) definite matrices. We train a single graph neural network to learn a mapping from an entire class of such matrices to prolongation operators, using an efficient unsupervised loss function.  Experiments on a broad class of problems demonstrate improved convergence rates compared to classical AMG, demonstrating the potential utility of neural networks for developing sparse system solvers. 

%% file: introduction.tex
Algebraic multigrid (AMG) is a well-developed efficient numerical approach for solving large ill-conditioned sparse linear systems and eigenproblems. Introduced in the 1980's \cite{BMR84,R83,RS87}, AMG and its many variants have been applied to diverse problems, including partial differential equations (PDEs), sparse Markov chains, and problems involving graph Laplacians, (e.g., \citet{AMGe,FreeAMGe,Heys05,stuben2001introduction,HL94,Vir06,Sterck07,Sterck08,TY2010,LEAN13,NN16,FM18}). While AMG is mathematically well grounded, its application involves the selection of problem-dependent parameters and heuristics, requiring expert knowledge and experience. Machine learning may therefore offer effective tools for developing {\it efficient} AMG algorithms. 

AMG is a multi-level iterative method for linear systems, 
\begin{equation}\label{eq:system}
Ax=b,
\end{equation}
where $A \in \Real^{n \times n}$ is a sparse matrix and $x,b \in \Real^n$, with $x$ the unknown solution vector. Given an initial approximate solution $x_0 \in \Real^n$, the $(k+1)$st iteration of AMG proceeds as follows, with details provided in Section \ref{problemDefinition}. Given the $k$th iteration, $x^{(k)}$, a few steps of a simple iterative solver (typically Gauss-Seidel relaxation) are applied, followed by the construction of a smaller linear system for the  error at a ``coarser scale''. This is done by selecting a subset of $n_c < n$ ``representative" variables (called the coarse variables), and constructing a \textit{prolongation} operator $P \in \Real^{n \times n_c}$ relating the coarse variables in $\Real^{n_c}$ to the variables in $\Real^n$. This smaller problem is treated recursively, by applying relaxation and appealing to a still coarser representation, and so on. Using $P$, the resulting solution is then ``prolongated'' back to the fine level to update the approximate solution, and a few additional relaxation sweeps are applied, yielding $x^{(k+1)}$. Note that the AMG procedure is analogous to the classical geometric multigrid algorithm (GMG) \cite{Bra77,BHM00,TOS01}, but unlike GMG (and other common multilevel algorithms) the variables need not lie on a regular grid or even be associated with a metric space.

The AMG procedure involves two critical heuristics which are applied at each level of the recursion, selection of coarse variables and construction of the prolongation matrix. Here we will use machine learning to address the latter heuristic. The choice of prolongation matrix $P$ critically depends on $A$, and it strongly influences the efficiency of the AMG algorithm. After decades of research which yielded numerous theoretical insights and practical developments, there is still no single recipe for constructing prolongation operators that are optimal for a given class of problems. 
This paper proposes a framework for learning maps from entire classes of sparse symmetric positive definite and semidefinite (SPD/SPSD) matrices to prolongation operators, yielding efficient AMG solvers. Given a class of sparse SPD/SPSD matrices (e.g., low-degree graph Laplacian operators whose entries are drawn from a given distribution), we train a single network to solve any linear system of equations with a matrix drawn from that class. To train our network, we first represent the matrix $A$ in \eqref{eq:system} by a graph $G_A=(V_A,E_A)$, where the vertex set $V_A$ contains a vertex per each variable $x_i$ and the edge set $E_A$ contains an edge $e_{ij}$, with a corresponding weight $A_{ij}$, if and only if $A_{ij}\neq 0$. Learning prolongation operators then becomes a graph learning problem which takes as input edge and node features and outputs edge weights on a subset of $E_A$.
Specifically, we utilize  a graph learning algorithm based on message passing \cite{gilmer2017neural, battaglia2018relational}. 

This paper generalizes the work of \citet{greenfeld2019learning}, which is restricted to 2D diffusion partial differential equations discretized on a rectangular grid, and therefore cannot be applied to unstructured problems. In contrast, AMG handles general sparse SPD/SPSD matrices, with varying node degree in the graph $G_A$.  Furthermore, development of new AMG approaches by experts is challenging, so the potential gain in using machine learning might be significant. Finally, in order to achieve efficient training, we introduce a novel Fourier analysis for locally unstructured problems, by constructing a block-periodic triangular mesh. Our experiments demonstrate the utility of our approach, showing in particular that our method can generalize across problem size, graph topology, and distribution, demonstrating better convergence rates than those achieved by classical AMG.

%% file: relatedWork.tex
Amongst recent papers on using machine learning for linear system solvers, our work most closely follows \citet{greenfeld2019learning}, which uses a multilayer perceptron (MLP) with skip-connections to produce prolongation operators for 2D diffusion partial differential equations discretized on a rectangular grid. As noted above, here we lift all such structure restrictions. To the best of our knowledge, our work is the first to apply neural networks for solving such broad classes of sparse linear systems.

Another notable related paper is \citet{hsieh2019learning}, which uses a convolutional network to improve on an existing linear iterative solver. In particular, learning is applied to improve a GMG algorithm for structured Poisson problems in an end-to-end manner, by using a U-Net architecture with several downsampling and upsampling layers, and learning from supervised data. \citet{schmitt2019optimizing} use evolutionary methods to optimize a GMG solver. \citet{katrutsa2017deep} optimize restriction and prolongation operators for GMG, by formulating the entire two-grid algorithm as a deep neural network, and approximately minimizing the spectral radius of the resulting iteration matrix. They evaluate their method on single instances of various structured-grid differential equations in 1D. \citet{sun2003solving} use a tailored network with a single hidden layer to solve the Poisson equation on a specific mesh.

\paragraph{Graph Neural Networks.}
Learning graph-structured data is an important and challenging learning setup that has received significant attention in recent years. The main challenge stems from the fact that graphs vary in size and topology, and also that graphs adhere to specific data symmetries (e.g., node reordering), which hinders the ability to use simple models such as MLPs. The first neural networks for graphs were proposed in \citet{gori2005new,Scarselli2009}. Since then, a plethora of architectures were proposed, which can be roughly divided into two types: (1) spectral-based methods (e.g., \citet{Bruna2013,Henaff2015,Defferrard2016}), that define graph convolutions as diagonal operators in the Graph Laplacian eigenbasis, and (2) \emph{message-passing neural networks} \cite{gilmer2017neural, battaglia2018relational}, which are currently the most popular and flexible architectures. In a nutshell, these models maintain a feature vector for each node in the graph, and update it by applying a parametric function (often an MLP) to the features of neighboring nodes.

Graph neural networks have been applied to various problems including molecule property prediction \cite{gilmer2017neural}, social network analysis \cite{kipf2016semi} and point-cloud and shape analysis \cite{wang2019dynamic}. Recently, several papers (e.g., \citet{selsam2018learning,li2018combinatorial}), targeted the task of solving combinatorial optimization problems efficiently using graph neural networks. Similarly to our work, their network is trained on small problems and is able to generalize to much larger problems, and to different distributions.

%% file: problemDefinition.tex
AMG algorithms employ a hierarchy of progressively coarser approximations to the linear system under consideration, to accelerate the convergence of classical simple and cheap iterative processes called {\em relaxation} (most commonly Gauss-Seidel). For the SPD/SPSD problems we are considering, relaxation is known to be efficient for reducing so-called high-energy error modes, that is, error comprised primarily of eigenvectors of $A$ with relatively large eigenvalues. On the other hand, relaxation is extremely inefficient for low-energy error comprised of eigenvectors with small corresponding eigenvalues \cite{RDF06}. The coarse-level correction, as described below, complements the relaxation by efficiently reducing low-energy modes, resulting in an efficient solver.

For a basic description of AMG, consider again the linear system $Ax=b \, ,$ where $A$ is a real sparse SPD matrix of size $n \times n$, $b \in \Real^n$, and $x \in \Real^n$ is the unknown solution vector. The two-level AMG algorithm is defined in Algorithm \ref{alg:TL}, with ``relaxation sweeps'' referring to iterations of the prescribed relaxation process, typically the classical Gauss-Seidel relaxation as defined below. For a detailed description of the two-level and multi-level AMG algorithm we refer the reader to classical textbooks \cite{BHM00,stuben2001introduction}. 

\begin{algorithm}
   \caption{Two-Level Algorithm}
   \label{alg:TL}
\begin{algorithmic}[1]
   \STATE {\bfseries Input:} SPD matrix $A \in {\mathbb R}^{n \times n}$, initial approximation $x^{(0)} \in {\mathbb R}^n$, right-hand side $b \in {\mathbb R}^n$, full-rank prolongation matrix $P \in {\mathbb R}^{n \times n_c}$,  a relaxation scheme, $k = 0$, residual tolerance $\delta$.
   \REPEAT
    \STATE Perform $s_1$ relaxation sweeps starting with the current approximation $x^{(k)}$, obtaining ${\tilde x}^{(k)}$.
    \STATE Compute the residual: $r^{(k)}=b-A {\tilde x}^{(k)}$.
    \STATE Project the error equations to the coarser level and solve the coarse-level system: $A_c e_c^{(k)} = P^T r^{(k)}$, with $A_c = P^T A P$.
    \STATE Prolongate and add the coarse-level solution: ${\tilde x}^{(k)}={\tilde x}^{(k)}+P e_c^{(k)}$.
    \STATE Perform $s_2$ relaxation sweeps obtaining $x^{(k+1)}$.
    \STATE $k = k+1$.
    \UNTIL{$||r^{(k-1)}||<\delta$}.
\end{algorithmic}
\end{algorithm}

\noindent
The prolongation $P$ in Algorithm \ref{alg:TL} is a sparse, full column-rank matrix, with $n_c < n$, and therefore $A_c$ is a sparse SPD matrix of size $n_c \times n_c$, hence smaller than $A$. This allows us to apply the algorithm recursively. That is, in the multi-level (or multigrid) version of the algorithm, the exact solution in Step 5 is replaced by one or more recursive calls to the two-level algorithm, employing successively coarser levels (smaller matrices). An iteration with a single recursive call is known as a \emph{V-cycle}, whereas an iteration with two calls is known as a \emph{W-cycle} (motivated by the shape of the recursive call tree). These recursive calls are repeated until reaching a very small problem, which is solved cheaply by relaxation or an exact solve. Thus, the multi-level AMG algorithm applies iterations until convergence, with each iteration employing the recursive structure as described.

It can be seen that the two-level AMG algorithm is comprised of two main components: the relaxation sweeps performed in Line 3 and Line 7 of the algorithm, and the coarse-level correction process described in Lines 4-6. For relaxation, we adopt Gauss-Seidel iteration, which is induced by the splitting $A = L + U$, where $L$ is the lower triangular part of $A$, including the diagonal, and $U$ is the strictly upper triangular part of $A$.
The resulting iterative scheme,
\begin{equation} \label{Gauss_Seidel}
x^{(m)} = x^{(m-1)} + L^{-1} \left(b-A x^{(m-1)} \right) \, ,
\end{equation}
which defines the relaxation sweeps in Line 3 and Line 7, is convergent for SPD matrices\footnote{The total number of arithmetic operations required for a single Gauss-Seidel relaxation sweep is roughly equal to the number of non-zero elements in $A$, assumed to be $O(n)$.}.
Here, $(m)$, the superscript, denotes the iteration number of the Gauss-Seidel relaxation.
The error after iteration $m$, $e^{(m)} = x - x^{(m)}$, is related to the error before the iteration by the error propagation equation,
\begin{equation} \label{Gauss_Seidel_Error_Propagation}
e^{(m)} = S e^{(m-1)} \, ,
\end{equation}
where $S = I - L^{-1}A$ is called the error propagation matrix of Gauss-Seidel relaxation, with $I$ denoting the identity matrix of the same dimension as $A$.

The error propagation equation of the entire two-level algorithm is given by
\begin{equation} \label{Two_Grid_Error_Propagation}
e^{(k)} = M e^{(k-1)},
\end{equation}

\noindent
where $M = M(A,P) = M(A,P;S,s_1,s_2)$ is the two-level error propagation matrix
\begin{equation}\label{eq:M}
M = S^{s_2} C  S^{s_1}.
\end{equation}

\noindent
Here, $s_1$ and $s_2$ are the number of relaxation sweeps performed before and after the coarse-level correction process, and $C$ is the error propagation matrix of the coarse-level correction, given by
\begin{equation}\label{eq:C}
C = I-P \left[ P ^T A P \right]^{-1} P^T A.
\end{equation}

\noindent
For a given operator $A$, the error propagation matrix $M$ defined in \eqref{eq:M} governs the convergence behavior of the two-level (and consequently multi-level) cycle. The key to designing effective AMG algorithms of this form lies in the selection of the prolongation matrix $P$. The relaxation and coarse-level correction process play complementary roles. That is, the solver may be efficient only if the error propagation matrix $C$ of the coarse-level correction process significantly reduces low energy errors, because $S$ only reduces efficiently high-energy errors, as noted above. Observe, on the other hand, that $CP = 0$, implying that the coarse-level correction eliminates any error that is in the subspace spanned by the columns of $P$. Indeed, the matrix $P \left[ P ^T A P \right]^{-1} P^T A$ in \eqref{eq:C} is an $A$-orthogonal projection onto the range of $P$. It follows that we must construct $P$ such that all low-energy errors will approximately be in its range. At the same time, $P$ also needs to be very sparse for computational efficiency.

\subsection{Constructing $P$}

AMG algorithms typically divide the task of constructing $P$ into three phases. 
\noindent
The first step is a partitioning of the nodes of the graph $G_A$ into  ``C-nodes" and ``F-nodes", where C and F stand for {\em coarse} and {\em fine}. The C-nodes comprise the ``coarse grid'', which is  a subset of the ``fine grid'' comprised of all the nodes. The partitioning is performed on the basis of the nonzero off-diagonal elements of $A$ (see, e.g., \citet{BHM00,stuben2001introduction}) for detailed examples). The resulting $n_c$ C-nodes correspond to the columns of $P$, while all the nodes of $A$ correspond to the rows. The second step is selecting the sparsity pattern of $P$, also based on the elements of $A$.
The final step is to select the values of the nonzero elements $P$. If row $i$ of $P$ corresponds to a C-point, say the one corresponding to column $j$, then $P_{i,j}$ is set to 1. The remaining nonzero values of $P$ are selected by formulas or processes depending locally on the elements of $A$, that is, on the $i$th row of $A$ and rows corresponding to nodes that are a short distance from node $i$ on the graph of $A$. 

In this paper we focus on the final step, the goal of selecting the nonzero values of $P$. To this end, we select the C-nodes and the sparsity pattern of $P$ (first and second steps) according to the well-known classical AMG (CAMG) algorithm, as implemented in \citet{OlSc2018}. Then, we  employ a learning process for deriving network-based formulas for the nonzero values of $P$ based locally on the elements of the matrix $A$. We then compare the resulting solver to classical AMG, demonstrating improved convergence rates. This suggests that machine learning methods can provide an improvement over formulas that have been developed by experts over decades of research. The details of the learning process are provided in the next section.

%% file: method.tex
\newcommand{\norm}[1]{\left\lVert#1\right\rVert}

Our task is to learn a mapping $P=P_\theta(A)$, where $A$ is a sparse square matrix, $\theta$ are the learned parameters, and $P$ is the resulting prolongation matrix. As discussed above, $P$ should satisfy two objectives: it should be very sparse, and the resulting two-level algorithm should yield fast convergence. The first objective is satisfied by imposing a sparsity pattern on $P$ derived from the classical AMG algorithm. For the second objective, the asymptotic convergence rate of the two-level algorithm is governed by the spectral radius of the error propagation matrix $M(A,P)$ \eqref{eq:M}, which we aim to approximately minimize.

Since backpropagation through Eigendecomposition tends to be numerically unstable \cite{wang2019backpropagation}, we relax the objective to the squared Frobenius norm, which bounds the spectral radius from above. Hence, given a distribution $\mathcal{D}$ over linear operators, $A$, for some fixed relaxation $S$ and parameters $s_1$ and $s_2$, we  define the following unsupervised learning problem 
\begin{equation} \label{eq:loss} 
\min_{\theta} \mathbf{E}_{A\sim\mathcal{D}}
\norm{(M(A,P_\theta(A)))}^2_F 
\end{equation}
where the data are only the elements of $A$, which are drawn from some distribution $\mathcal{D}$.
\subsection{Learning the prolongation operator} \label{Graph_Learning}

As explained in the introduction, the linear system $A$ is represented as a graph $G_A=(V_A,E_A)$, with nodes corresponding to the variables, and edges corresponding to non-zero elements of $A$. Therefore, the problem of setting values to the prolongation matrix $P$ amounts to assigning a set of values $\{p_e \}_{e\in E_A^c}$, where $E^c_A \subset E_A$ is defined according to the given sparsity pattern, i.e., a set of edges that connect C-nodes to F-nodes (and to themselves). See illustration of a small problem  in  Figure \ref{fig:edges}. Using this formalism, the task of selecting the prolongation weights can naturally be formulated as a graph learning problem: given the matrix $A$, a set of node features $\{f_v\}_{v\in V_A}$ and a set of edge features $\{f_e \}_{e\in E_A}$, we construct the graph $G_A$ and use a graph neural network 
$$P_\theta \left(G_A,\{f_v\}_{v\in V_A},\{f_e \}_{e\in E_A}\right)$$
to predict the prolongation weights $\{p_e \}_{e\in E_A^c}$. In our case, the vertex features indicate whether the vertex is a C-point or not, and the edge features are comprised of the edge weights $A_{ij}$ as well as the indicator of $E_A^c$ that represents the sparsity pattern.
As a final step, we scale each row of $P$ to have the same row sum as the prolongation produced by the classical AMG algorithm\footnote{The important task of learning the optimal scaling is left to future research.}. The resulting prolongation operator is not guaranteed to be a full-rank matrix, but since singular matrices result in high loss, in our experiments the trained networks produced only full-rank matrices.
\begin{figure}[h]
    \centering
         \includegraphics[width=\columnwidth]{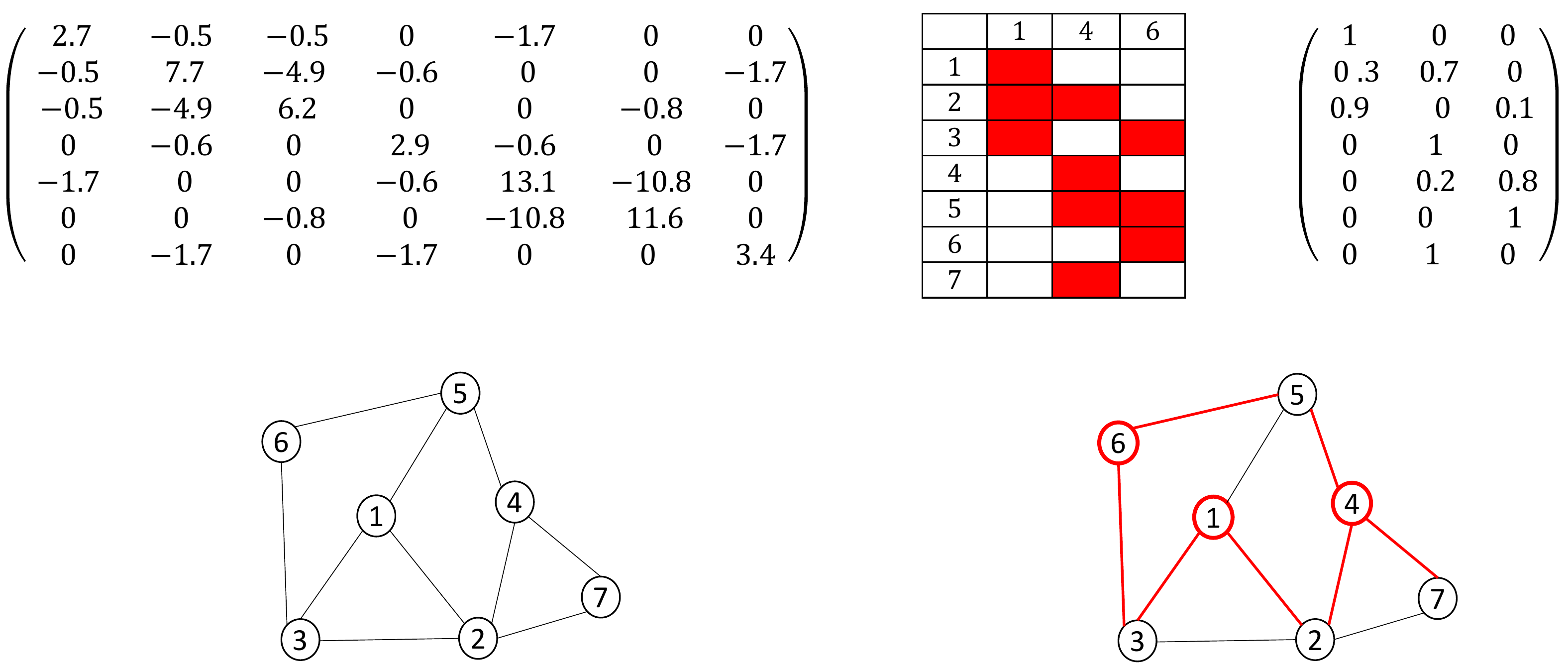}
         \label{fig:edges}
    \caption{Example of sparse matrix $A$ (upper left) and its associated graph $G_A$ (lower left). The sparsity pattern (upper middle) defines the set of C-nodes ${1,4,6}$, and to which nodes each C-node contributes (denoted by the red cells). In the graph at the lower right panel, red edges represent the set $E^c_A$ corresponding to the sparsity pattern. At the upper right is the prolongation matrix $P$, which is formed by setting weights to the sparsity pattern. Self-edges are omitted, for clarity.}
    \label{fig:edges}
\end{figure}

\subsection{Network Architecture} \label{Network_Architecture}
\paragraph{Layers.} Three main considerations come up when choosing a concrete GNN architecture that includes appropriate layers for our problem: (1) efficiency: the run-time of the mapping from $A$ to $P$ should be proportional to the number of nonzero elements, $O(n)$; (2) flexibility: the architecture should be able to process graphs of different size and connectivity; (3) edge-features: ability to process and output edge features. The first requirement rules out recently suggested layers as in \citet{maron2019provably,chen2019equivalence}, which suffer from higher complexity, while the second requirement rules out spectral methods (e.g., \citet{Bruna2013,Henaff2015,Defferrard2016}). One type of layer that does fulfill all these requirements is the layer suggested in the Graph Network (GN) framework of \citet{battaglia2018relational}, which generalizes many message passing variants and extends them to allow using edge features. Each such layer is comprised of two steps: a vertex feature update step and an edge feature update step. Each of these steps is implemented by a parameterized update function (an MLP) and a summation operation for aggregating multiple neighboring features into a single feature vector.

\paragraph{Architecture.} We use a variant of the encode-process-decode architecture suggested in \citet{battaglia2018relational}. This architecture is composed of three main parts: (1) an encoder followed by (2) a message-passing block and finally (3) a decoder. The encoder applies an MLP to the input features resulting in features of dimension 64. The message-passing block\footnote{Because the message passing architecture we use applies only to directed graphs, we represent the symmetric matrix $A$ as a directed graph with a pair of anti-parallel edges if two nodes are connected.} is composed of three message passing layers, each of which receives as input the output of the previous layer, concatenated with the encoder features. This is intended to allow each message passing round to efficiently utilize the edge weights, the coarse nodes and sparsity pattern information. Finally, an MLP decoder independently maps each edge feature to a feature of size one that represents the prolongation weight. All MLPs have four layers of width 64, and apply ReLU activation.

For efficiency reasons, existing AMG algorithms derive the prolongation weights from local information. Similarly, we use a small number of message passing rounds, so the prediction on each edge is a function of edges only a few hops away. For a bounded-degree graph, the run-time of each message-passing round, for the entire graph, is proportional to the size of the graph, therefore we achieve the required $O(n)$ run-time. Moreover, the local nature of the computation allows the network to learn rules for constructing prolongation operators of arbitrary size, as is demonstrated in the experiments section.

\subsection{Efficient Training on Block-Circulant Matrices}
\label{sec:efficient_training}

Our network is able to generalize to problems considerably larger than the problems it saw during training, but moderately large problems are still required for training. The main computational bottleneck when training the network is the computation of the error propagation matrix $M$ \eqref{eq:M}, which involves inversion of the coarse-level matrix $P^TAP$ of size $n^c \times n^c$, where $n^c$ is the number of nodes in the coarse-level graph. The cost of inverting a matrix may be as high as $O(n^3)$, because $n/n^c = O(1)$. The cost of other computations in training is $O(n)$ if implemented efficiently\footnote{Since the automatic differentiation software we use does not have complete support for sparse matrix operations, these computations have cost $O(n^2)$. In practice, this has not been a bottleneck in our work.}, therefore, for large problems the run-time of each training step is dominated by the inversion of $P^{T}AP$.

Generalizing to unstructured problems an approach used in \citet{greenfeld2019learning}, we reduce the training complexity by training on a limited class of $A$ matrices called block-circulant matrices. A block-circulant matrix $A$ of size $n\times n$, with $n=kb$, takes the form
\begin{equation*}
A=\left(\begin{array}{c}
A^{\left(0\right)}\\
A^{\left(1\right)}\\
A^{\left(2\right)}\\
\vdots\\
A^{\left(b-1\right)}
\end{array}\right),
\end{equation*}
where the blocks $A^{\left(m\right)},m=0,\dots,b-1$ are $k\times n$ submatrices whose elements satisfy
\begin{equation} \label{block_property}
    A^{\left(m\right)}_{l,j} = A_{l,\mod\left(j-k,n\right)}^{\left(m-1\right)}, \  m=1,\dots,b-1,
\end{equation}
and hence $A_{l,j}=A_{\mod(l-k,n),\mod(j-k,n)}$, where mod$(x,y)$ is the remainder obtained when dividing integer $x$ by integer $y$. In \citet{greenfeld2019learning} and the associated supplementary material, it is proved that such matrices are unitarily block-diagonalized by an appropriate Fourier basis. Furthermore, because the graphs associated with the matrices we use for training (here as well as in \citet{greenfeld2019learning}) are doubly block-periodic in the plane, each of the $b$ blocks of size $k$ by $k$ in the matrix resulting from the block-diagonalization is itself block-circulant, comprised of $b$ blocks of size $c$, with $k = bc$. The upshot is that the matrix $A$ can be unitarily transformed into a similar matrix that is block-diagonal with $b^2$ blocks of size $c$. Thus, if we wish to compute the spectral radius or Frobenius norm of $A$, we can compute these values for the block-diagonal matrix, requiring us to process $b^2$ matrices of size $c$ by $c$ rather than a large matrix of size $n$ by $n$, with $n = b^2c$. Since the block diagonalization itself is done analytically using the Fourier basis, it is cheap, and the overall cost of the entire computation is just linear in $n$ (assuming $c$ is a constant independent of $n$). To use this approach for training, we must make sure that $M$ in \eqref{eq:M} inherits the block-circulant form of $A$ (with a smaller value of $c$ due to the coarsening). We explain how we ensure this below.         
%As explained in detail in \citet{greenfeld2019learning}, this allows us to replace the direct processing of the resulting block-circulant $M$ in \eqref{eq:M} by the processing of multiple matrices of constant size. 

\begin{figure}
    \centering
    \includegraphics[width=0.3\columnwidth]{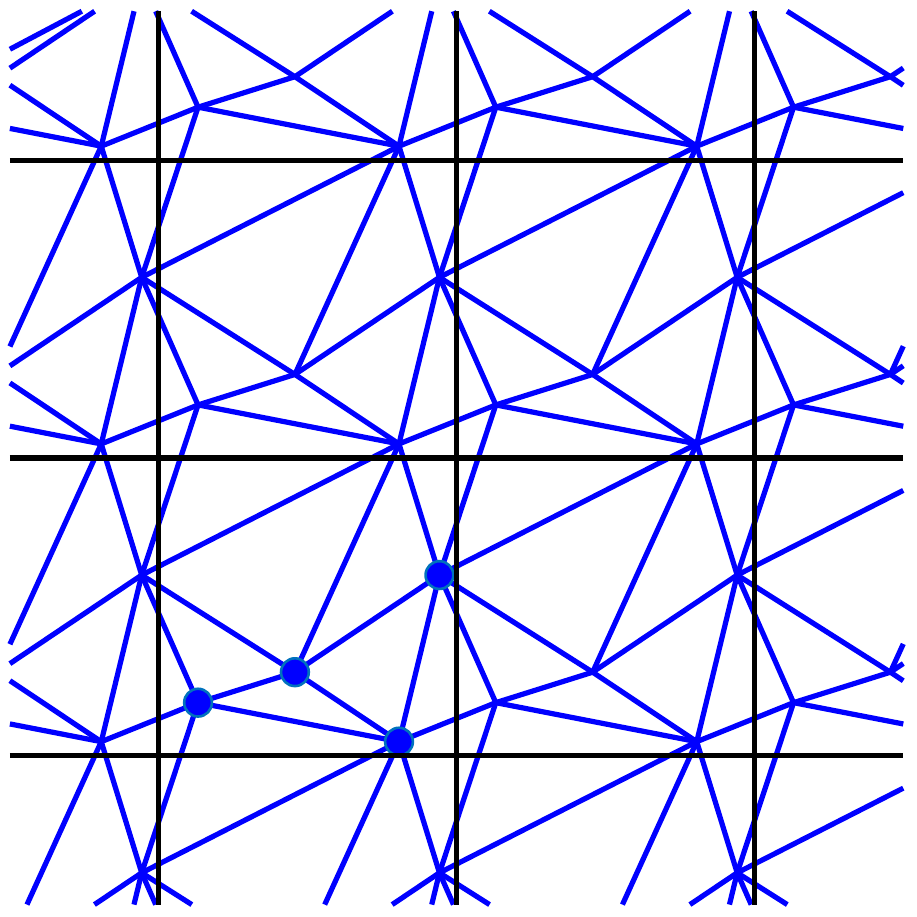}
    \caption{Example of a portion of a block-periodic Delaunay mesh with blocks of size $c=4$}
    \label{fig:periodic_delaunay}
\end{figure}

To create a locally unstructured block-circulant matrix $A$, we select $c$ random points on a square, and tile a large square domain with $b$ by $b$ such identical blocks. Now we apply Delaunay triangulation in the entire domain, and modify the edges near the boundaries of the domain so as to impose periodicity. Figure \ref{fig:periodic_delaunay} depicts a small portion of such a graph. Next, we number the nodes consistently, such that the $c$ nodes within each block are ordered contiguously and with the same ordering in all the blocks, while the blocks are ordered by the standard column-first ordering. Finally, we randomly select edge-weights for a single block according to the prescribed distribution, and replicate them to all the blocks. We thus obtain a graph whose Laplacian $A$ of size $b^2c$ by $b^2c$ is block-circulant as explained above, and can be transformed into a similar matrix that is block-diagonal with $b^2$ blocks of size $c$ by $c$.

To ensure that $M$ inherits the block-circulant structure of $A$ (with smaller $c$ as mentioned above), we must impose that the prolongation $P$ and relaxation $S$ have the same block-circulant form as $A$. (For this statement to be formally well-defined, we must make $P$ square by inserting a column of zeros per each $F$-node, but this has no influence on $M$.)   
%The prolongation matrix $P$ created by the AMG algorithm is of size $n\times n_{c}$, where $n_{c}$ is the number of coarse nodes. However, because the graph corresponding to $P$ has the same $n$ nodes as $G_A$, $P$ can be written as a square matrix\footnote{In the square form of the matrix $P$, the columns corresponding to F-nodes contain only zeros} of size $n\times n$. 
Even though $A$ is block-circulant, the matrix $P$ (in its square form) is not a priori guaranteed to be block-circulant, but in practice, we found that for the standard algorithms it is very close to block-circulant. To make it exactly block-circulant, we choose the block with the most common sparsity pattern, and tile $P$ with it. The $S$ matrix corresponding to Gauss-Seidel relaxation on the block-circulant matrix $A$, is block-circulant (for the bounded-degree graphs we are considering) only in the limit of infinite $n$. Nevertheless, the approximation of treating it as such for finite graphs by the Fourier analysis, as is commonly done in standard multigrid Fourier analysis, does not unduly affect performance in our experiments.

Finally, we remark on another advantage of the block Fourier analysis. In strictly positive semidefinite problems, such as the graph Laplacian, the matrix $P^T A P$ is singular, so $M$ in \eqref{eq:M} is undefined. The block-diagonalization allows us to isolate the single singular block and simply ignore it, and thus we do not need to artificially force $A$ to be nonsingular by adding a positive diagonal term.

%% file: experiments.tex
We compare the performance of our network based solver\footnote{Code for reproducing experiments is available at \url{https://github.com/ilayluz/learning-amg}.} to the well-known classical AMG (CAMG) algorithm of \citet{ruge1987algebraic}, as implemented in \citet{OlSc2018}. We evaluate performance by measuring the number of iterations (V-cycles or W-cycles) required to reach a specified accuracy and by estimating the asymptotic convergence factor per iteration (often called cycle). 

We focus on two different tasks: solving linear systems associated with graph Laplacian matrices with a variety of topologies, and solving diffusion partial differential equations discretized by linear finite elements over triangulated domains. Although the network is trained on a limited class of operators, namely block-circulant Laplacian matrices of relatively small size, where the coefficients are drawn from a lognormal distribution, it is able to generalize to larger problems, with diverse structure and distribution. This indicates that our network learns effective \emph{rules} for constructing prolongation operators, not just solvers for specific problems, due to the local nature of the computation. In addition, we test our network based solver in the role of a preconditioner in spectral clustering applications.

\paragraph{Input and output representation.}
As discussed above, the input to the network is a graph $G_A=(V_A,E_A)$ with a set of node features $\{f_v\}_{v\in V_A}$ and a set of edge features $\{f_e \}_{e\in E_A}$. The output is a set of scalar prolongation weights $\{p_e \}_{e\in E_A^c}$, where $E_A^c$ is defined by the given prolongation sparsity pattern. We represent node features by a one-hot encoding designating whether the node is a C-node $$f_v = \begin{cases} [1,0] &\mbox{if } v~~ \mbox{is a C-node} \\
[0,1] & \mbox{if } v~~ \mbox{is not a C-node} \end{cases}.$$
We represent edge features by a concatenation of the non-zero element of $A$ that corresponds to it, and a one-hot encoding designating whether the edge is part of the prolongation sparsity pattern $$f_e = \begin{cases} [A_{ij},1,0] &\mbox{if } e \in E_A^c \\
[A_{ij}, 0,1] & \mbox{if } e \not \in E_A^c \end{cases}.$$

\paragraph{Basis for comparison.}
The  algorithm we use for comparison, and for setting the sparsity pattern and row sum of the prolongation operator, is the CAMG algorithm \cite{ruge1987algebraic}, implemented in PyAMG \cite{OlSc2018}. For the selection of the coarse nodes, we use the strategy of CLJP  \cite{cleary1998coarse, alber2007parallel}, which selects a denser set of nodes than the default Ruge-Stuben algorithm \cite{ruge1987algebraic}. As is demonstrated in  Table \ref{tab:CAMG_comparison}, the CLJP algorithm has better asymptotic convergence rates on graph Laplacian problems than other C-node selection algorithms implemented in PyAMG, including Ruge-Stuben, Smoothed Aggregation \cite{SAV96}, Root-node Aggregation \cite{olson2011general}, and PMIS \cite{Sterck06}.
We use Gauss-Seidel relaxation, with $s_{1},s_{2}=1$. Because we use the same parameters in our method and the CAMG algorithm to which we compare, the run-time per iteration of the two algorithms is essentially the same. Of course, our setup time (which is applied once per test instance) is more expensive, because CAMG uses explicit formulas for computing the nonzero elements of $P$, whereas we use the trained network.  

\paragraph{Training details.}
The training data are comprised of block-circulant graph Laplacian matrices, composed of $4 \times 4$ blocks with 64 points in each block, yielding 1024 variables. The construction of such matrices follows the description in Sec. \ref{sec:efficient_training}, where the weights on the edges are drawn from standard lognormal distribution.  The network is trained to minimize the Frobenius norm of the two-level error propagation matrix $M$ in \eqref{eq:M}. In similar spirit as \citet{greenfeld2019learning}, the training is performed in two stages, first on the original problems and then on a training set comprised of the original problems and the once-coarsened problems as elaborated below. 
 
At the first stage we train on 256000 problems with $4\times 4$ blocks of size 64, with a single epoch. Then, we generate 128000 problems of $4\times 4$ blocks of size 128 and we apply the trained network to generate prolongation operators for each of those problems, and compute the block-circulant coarse matrices $A_{c}=P^{T}AP$. 
The CLJP C-node selection algorithm \cite{cleary1998coarse, alber2007parallel} selects roughly half of the nodes, so the coarsened problems are of approximately the same size as the original problems. We then generate 128000 additional problems with $4\times 4$ blocks of size 64, shuffle them with the coarsened problems, and continue training the network on the combined set of 256000 problems for another epoch\footnote{We may continue this process by training on twice-coarsened problems and so on. In practice however, we found that a network trained on a mixture of the original problem and the once-coarsened problem achieves good results even for large problems with multiple coarsening levels}. 

All experiments were conducted using the TensorFlow framework \cite{abadi2016tensorflow} using NVIDIA V100 GPU. We use a batch size of 32 and employ the Adam optimizer \cite{kingma2014adam} with a learning rate of $3\times 10^{-3}$.  Training took roughly 12 hours for first phase, another 12 hours for second phase. 

\subsection{Evaluation}

\begin{figure*}[t]
    \centering
    \begin{subfigure}[t]{0.32\linewidth}
        \centering
        \includegraphics[width=\linewidth]{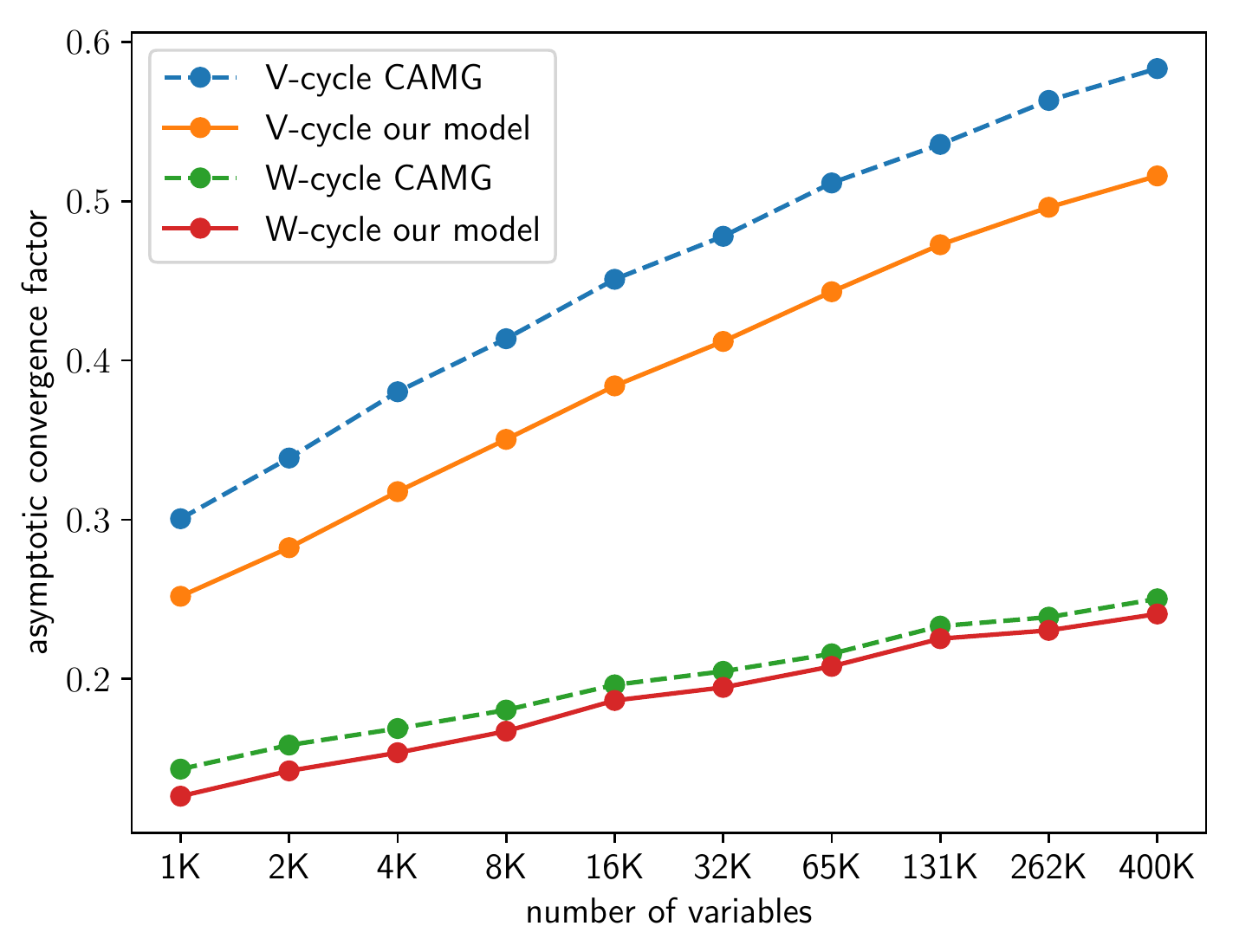}
        \caption{Graph Laplacian problems with lognormal distribution.}
        \label{fig:multilevel_lognormal_periodic}
    \end{subfigure}
    \hfill
    \begin{subfigure}[t]{0.32\linewidth}
        \centering
        \includegraphics[width=\linewidth]{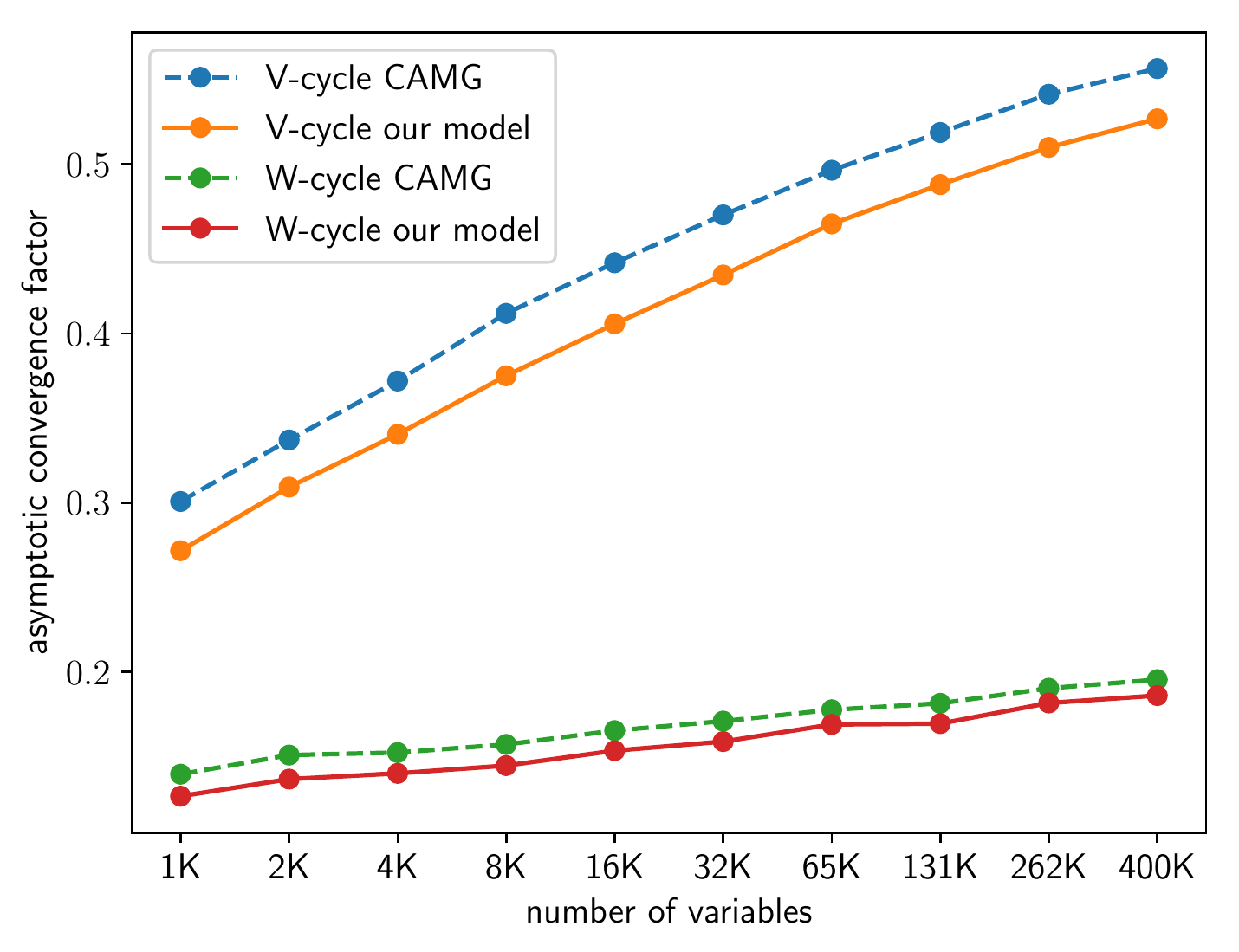}
        \caption{Graph Laplacian problems with uniform distribution.}
        \label{fig:multilevel_uniform_periodic}
    \end{subfigure}
    \hfill
    \begin{subfigure}[t]{0.32\linewidth}
        \centering
        \includegraphics[width=\linewidth]{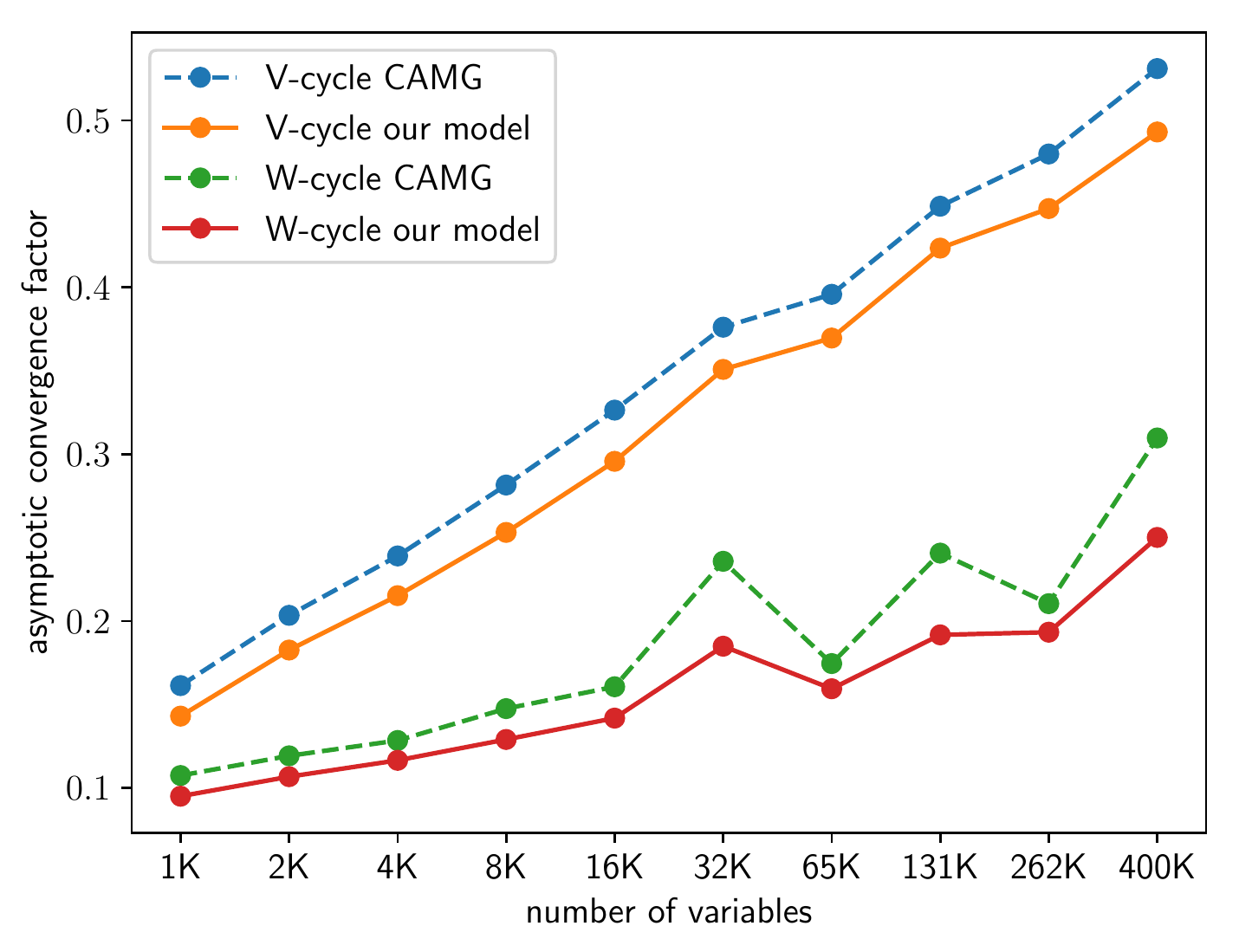}
        \caption{FEM diffusion equations.}
        \label{fig:multilevel_fe_complex_lognormal}
    \end{subfigure}
    \caption{Asymptotic convergence factors (smaller is better) for various problems. Each problem is tested on V and W-cycle, and averaged over 100 runs for each problem size}
    \label{fig:asymptotic_residual_plots}
\end{figure*}

\paragraph{Graph Laplacians.}
We first evaluate the performance of our network based solver on random graph Laplacian problems. To this end, we sample points uniformly on the unit square, and compute a Delaunay triangulation. Each edge is then given by a random weight sampled from a standard lognormal distribution, and the corresponding graph Laplacian matrix is constructed. 
We perform experiments on a range of problem sizes, with both V-cycles and W-cycles. We measure  the {\it asymptotic convergence factor} per cycle by initializing with a random $x^{(0)}$, performing 80 AMG cycles on the homogeneous problem\footnote{The asymptotic convergence factor is independent of the right-hand side $b$, so long as $b$ is in the range of $A$, i.e., has zero mean. We use $b=0$ so that we can perform many iterations without encountering roundoff errors (so long as we subtract off the mean so that the exact solution is zero), allowing us to measure accurately the asymptotic factor.} $Ax=0$, and computing the ratio of the residual norms of the last two iterations, $\frac{||r^{(k+1)}||_2}{||r^{(k)}||_2}$. For W-cycles, this value is almost equal to the spectral radius of the error iteration matrix $M$. Figure \ref{fig:multilevel_lognormal_periodic} shows the asymptotic convergence factor on problem sizes ranging from 1024 to 400000, for CAMG and for our model. Table \ref{tab:multilevel_success_rate_periodic} shows the success rate of the network, defined as the percentage of problems where our model outperformed CAMG. Figure \ref{fig:multilevel_uniform_periodic} shows the asymptotic convergence factor for graph Laplacian problems where the edge weights are sampled from a uniform $U(0,1)$ distribution, rather than the lognormal distribution used in training. The results indicate that the network based solver performs better than CAMG, and generalizes to large problems and other distributions, structure and topology.  

\begin{table}[h]\scriptsize
    \centering
    \caption{Asymptotic convergence factors for graph Laplacian problem with lognormal distributions of size 65536, for heuristic CAMG solvers. Tested on W-cycle, averaged over 100 runs for each C-node selection algorithm}
    \vskip 0.15in
    \begin{tabular}{l|c}
        \toprule
        C-node algorithm & average convergence factor \\
        \midrule
        CLJP                  &  0.21 \\
        Ruge-Stuben           &  0.24 \\
        Smoothed Aggregation  &  0.68 \\
        Root-node Aggregation &  0.70 \\
        PMIS                  &  0.98 \\
        \bottomrule
    \end{tabular}
    \label{tab:CAMG_comparison}
\end{table}

\begin{table}[h]\scriptsize
    \centering
    \caption{Success rate measured for graph Laplacian problems with lognormal (columns 2,3) and uniform (columns 4,5) distributions. Tested on V- and W-cycles, averaged over 100 runs for each problem size}
    \vskip 0.15in
    \begin{tabular}{l|c|c||c|c}
        \toprule
        size & V-cycle & W-cycle & V-cycle & W-cycle\\
        \midrule
        1024   &  97\% & 83\%  &  83\% & 83\%\\
        2048   &  98\% & 91\%  &  84\% & 85\% \\
        4096   &  98\% & 91\%  &  84\% & 84\% \\
        8192   &  99\% & 84\%  &  91\% & 84\% \\
        16384  &  99\% & 79\%  &  92\% & 80\% \\
        32768  &  98\% & 78\%  &  89\% & 81\% \\
        65536  & 100\% & 79\%  &  88\% & 80\% \\
        131072 & 100\% & 76\%  &  91\% & 82\% \\
        262144 & 100\% & 83\%  &  94\% & 72\% \\
        400000 &  98\% & 82\% &  93\% & 78\% \\
        \bottomrule
    \end{tabular}
    \label{tab:multilevel_success_rate_periodic}
\end{table}

\paragraph{Diffusion equations.} We test the network based solver on a variety of diffusion partial differential equations, 
\begin{equation}\label{diffusion}
-\nabla\cdot({\bf g}\nabla {\bf u}) = {\bf f},
\end{equation}
discretized on 2D triangular meshes. Given a 2D triangular mesh, for each triangle we randomly select a positive diffusion coefficient and construct the corresponding linear system, using linear finite elements (FEM). The mesh is generated using the Triangle mesh generation software of \citet{shewchuk1996triangle}. The diffusion coefficients ${\bf g}_i$ are sampled from a lognormal distribution with a log-mean of zero and log-standard deviation of 0.5. Finally, we modify the operator at the boundaries to impose Dirichlet boundary conditions. The resulting matrix $A$ is SPD. 

\begin{figure}[h]
    \centering
         \includegraphics[width=0.3\columnwidth]{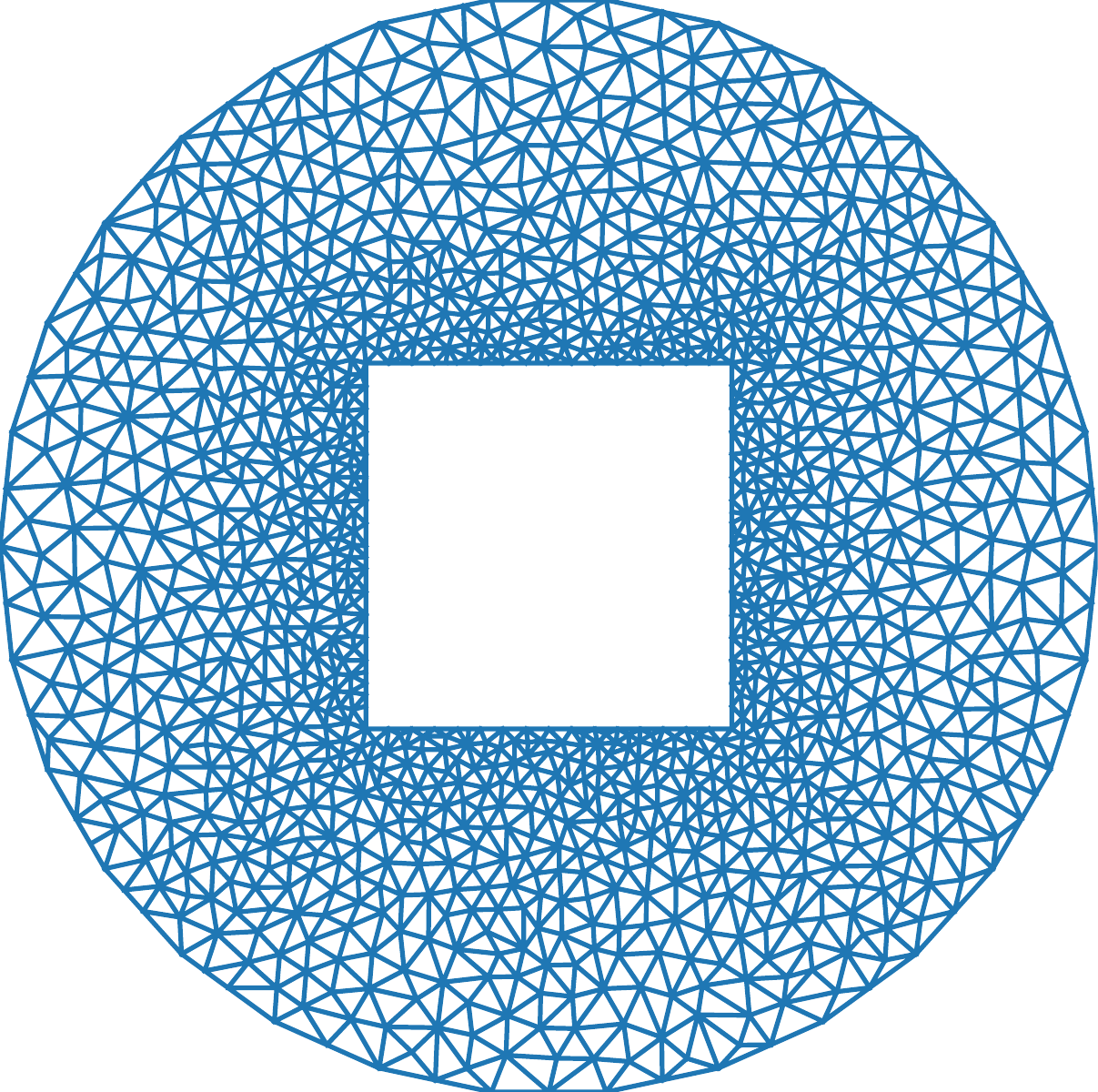}
         \label{fig:complex_fem_mesh}
    \caption{Example of a FEM mesh}
    \label{fig:fem_meshes}
\end{figure}

We test the same trained network as in the graph Laplacian problem (without any additional training) on a circular domain with a square hole and variable triangle density (see Figure \ref{fig:fem_meshes}).  

Figure \ref{fig:multilevel_fe_complex_lognormal} shows the asymptotic convergence factor for problem sizes ranging from 1024 to 400000, for CAMG and for the network based solver, averaged over 100 runs.  Table \ref{tab:multilevel_success_rate_fe_complex_lognormal} shows the success rate of the network, defined as the percentage of problems where our model outperforms CAMG. 

\begin{table}[]\scriptsize
    \centering
    \caption{Success rate measured for FEM diffusion equations. Tested on V and W-cycle, averaged over 100 runs for each problem size}
    \vskip 0.15in
    \begin{tabular}{l|c|c}
        \toprule
        size & V-cycle & W-cycle \\
        \midrule
        1024   &  87\% & 88\% \\
        2048   &  94\% & 85\% \\
        4096   &  99\% & 84\% \\
        8192   &  99\% & 90\% \\
        16384  &  96\% & 88\% \\
        32768  &  96\% & 96\% \\
        65536  &  98\% & 87\% \\
        131072 &  96\% & 94\% \\
        262144 &  97\% & 77\% \\
        400000 &  96\% & 89\% \\
        \bottomrule
    \end{tabular}
    \label{tab:multilevel_success_rate_fe_complex_lognormal}
\end{table}

\paragraph{Spectral Clustering.}
Spectral clustering is a widely used clustering algorithm \cite{von2007tutorial}. It involves computing eigenvectors associated with the smallest nonzero eigenvalues of a Laplacian matrix $A$ derived from a pairwise similarity measure of the data, and then performing a standard clustering algorithm (e.g., $k$-means) on them. In the case of large-scale sparse problems, these eigenvalues can be efficiently computed by an iterative preconditioned conjugate gradient method, such as LOBPCG \cite{knyazev2001toward} used in the popular Scikit-learn library \cite{scikit-learn}. At each iteration $i$, a matrix-vector product of the pseudo-inverse of $A$ and a residual vector $r_i$, i.e.,  $A^{\dagger}r_i$, is approximately computed by applying CAMG as a pre-conditioner to estimate  the solution of the linear system $Ax=r_i$.

\begin{figure}[h]
    \centering
     \begin{subfigure}[b]{0.3\columnwidth}
         \centering
         \includegraphics[width=\columnwidth]{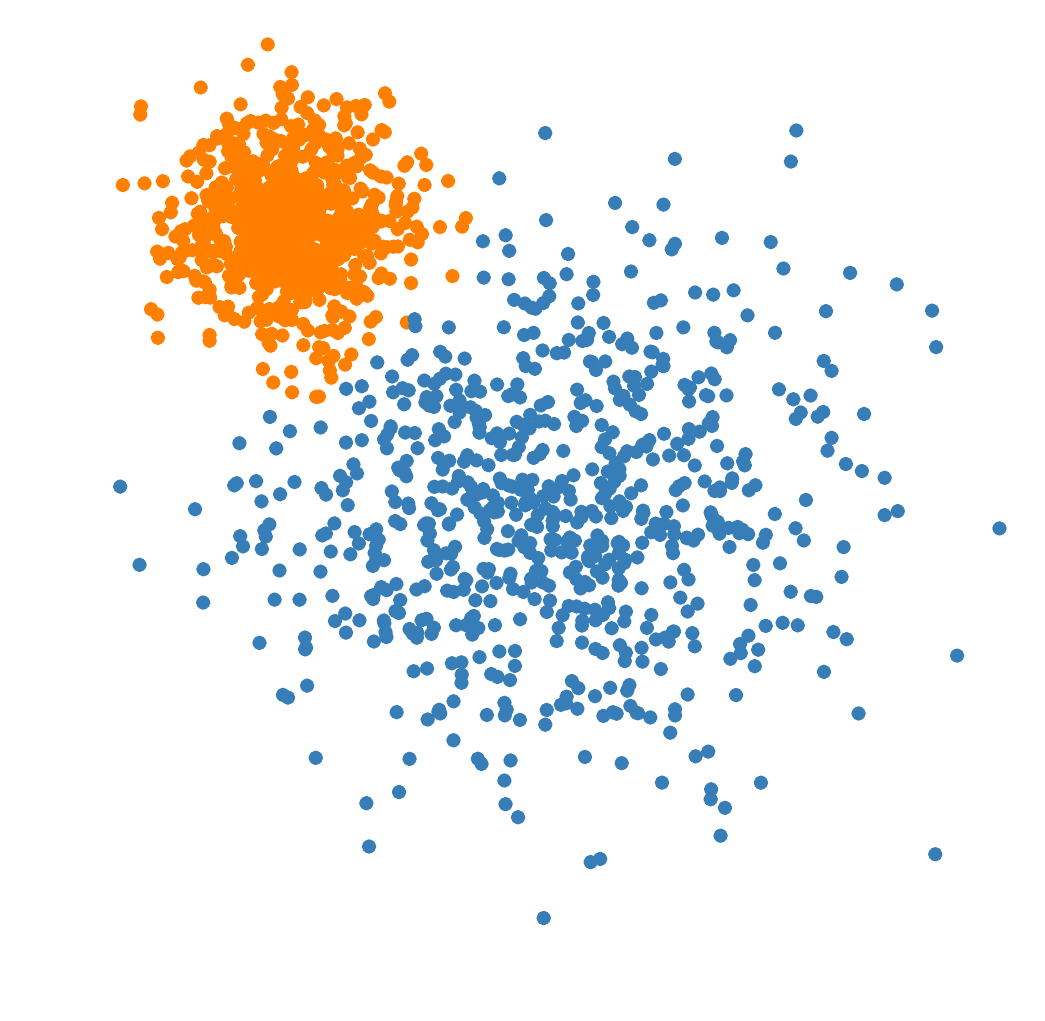}
         \caption{2 Gaussians}
         \label{fig:2_blobs}
     \end{subfigure}
     \qquad
     \begin{subfigure}[b]{0.3\columnwidth}
         \centering
         \includegraphics[width=\columnwidth]{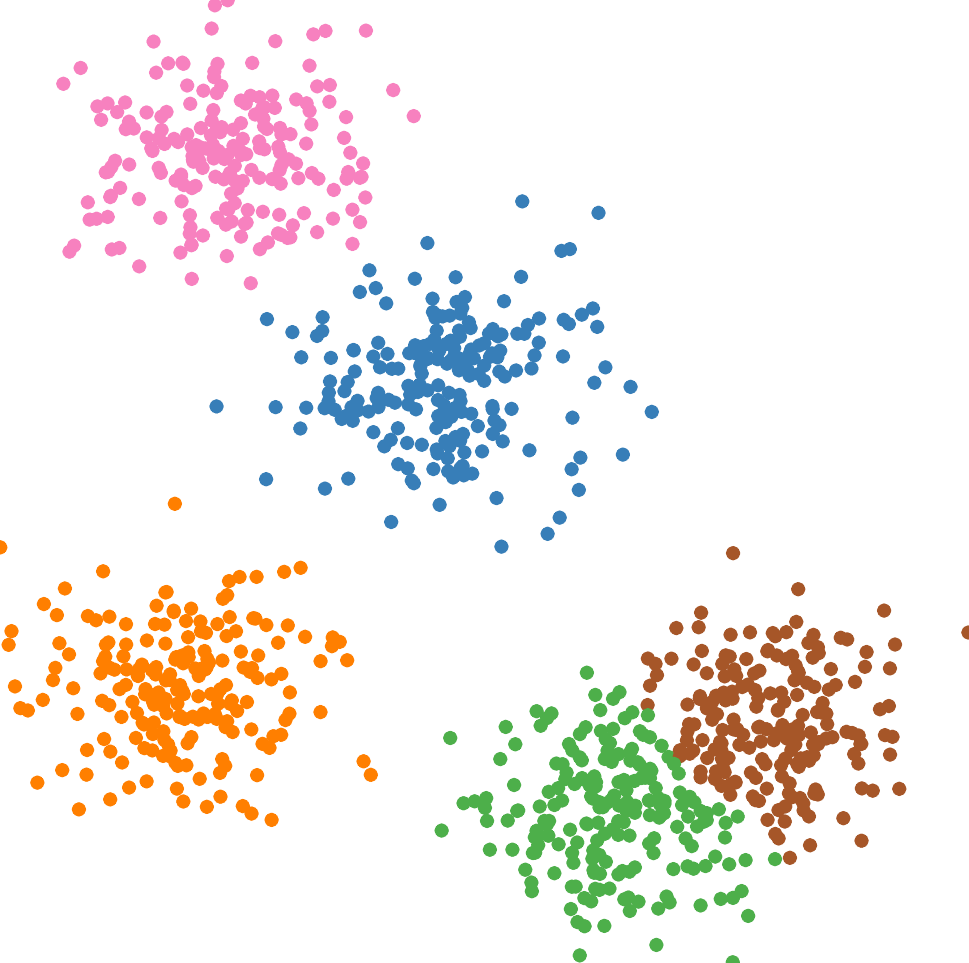}
         \caption{5 Gaussians}
         \label{fig:5_blobs}
     \end{subfigure}
     \vskip\baselineskip
     \begin{subfigure}[b]{0.3\columnwidth}
         \centering
         \includegraphics[width=\columnwidth]{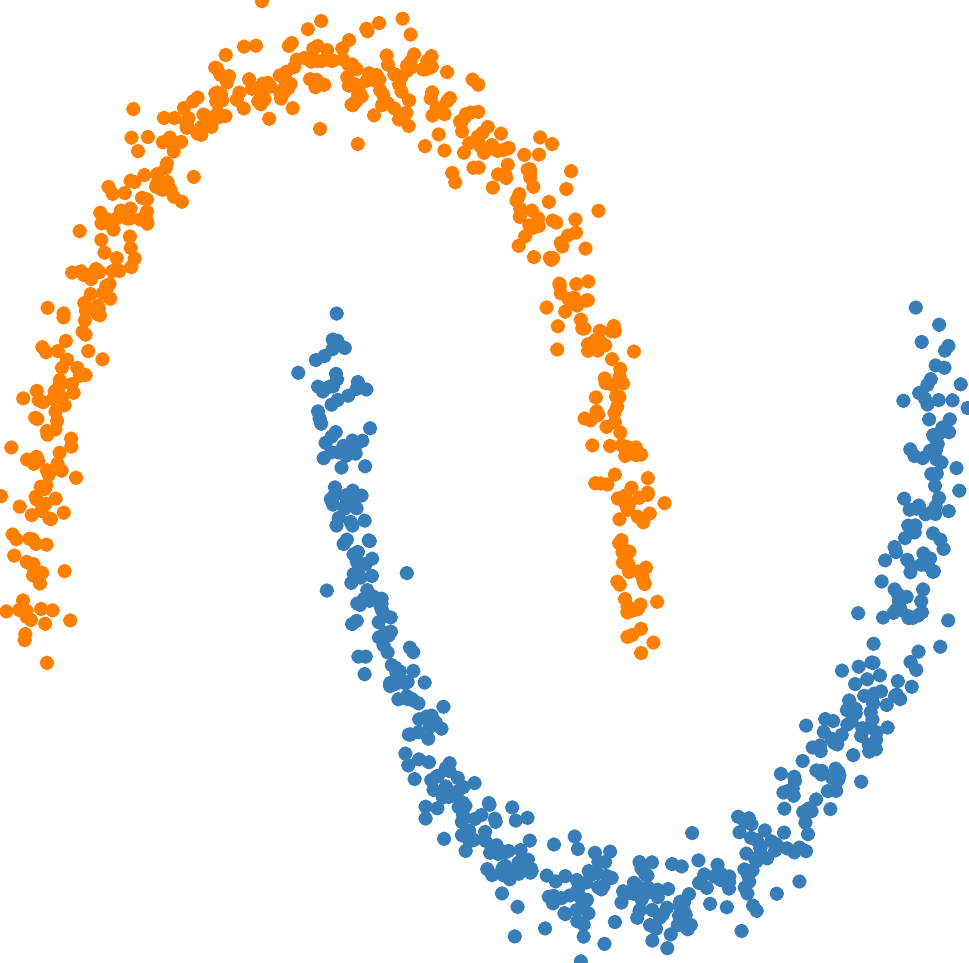}
         \caption{Moons}
         \label{fig:moons}
     \end{subfigure}
     \qquad
     \begin{subfigure}[b]{0.3\columnwidth}
         \centering
         \includegraphics[width=\columnwidth]{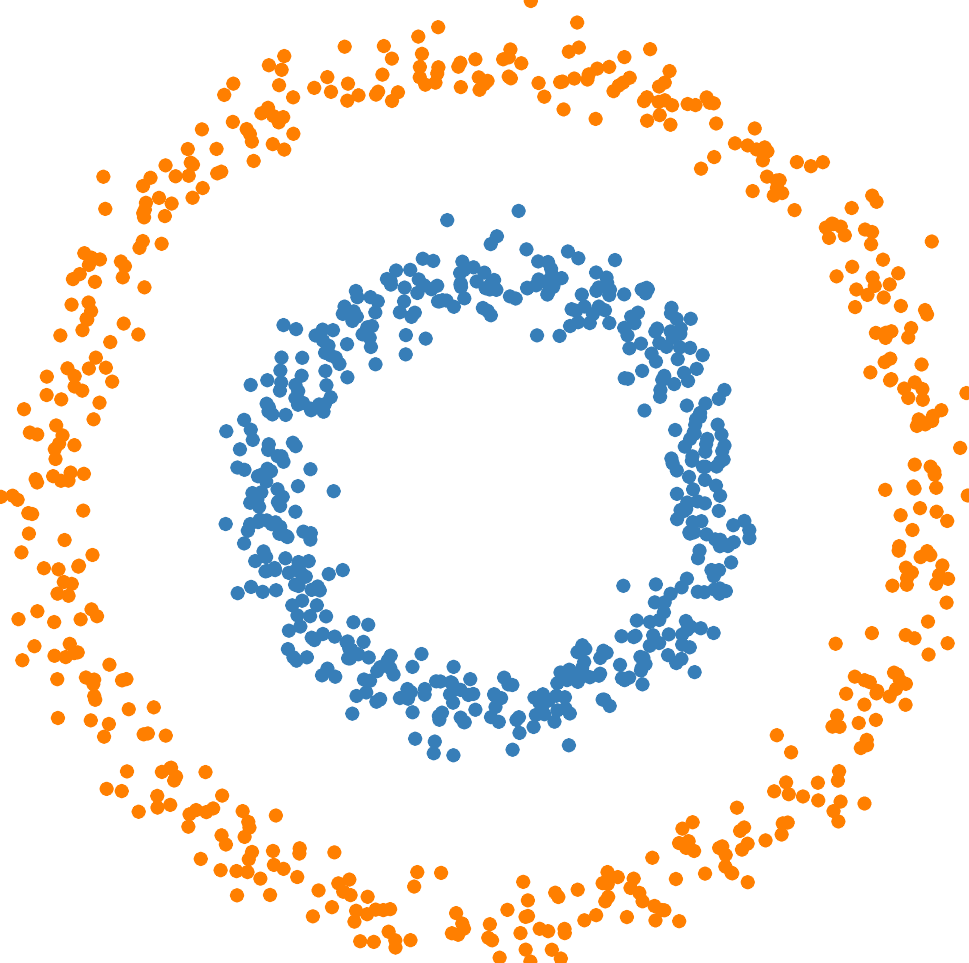}
         \caption{Circles}
         \label{fig:circles}
     \end{subfigure}
    \caption{Examples of our results for two dimensional spectral clustering problems}
    \label{fig:spec_examples}
\end{figure}

We evaluate the efficiency of our network based solver as a preconditioner by estimating the number of iterations needed to converge to a certain accuracy, and comparing with the CAMG preconditioner. To this end, we train our network with the same hyper-parameters and  generate training data as follows. Each training problem is produced by 1024  points sampled from two dimensional isotropic Gaussian distributions, one with standard deviation 1.0, the other with standard deviation 2.5, and the two centers are uniformly sampled from $[-10, 10]^2$ (see Figure \ref{fig:2_blobs}, for example). We compute the Euclidean $k$-nearest neighbors for $k=10$, and convert the distances to affinity measures by setting $S_{ij}=e^{-d_{ij}^2}$, where $d_{ij}$ is the distance between two different points $i$ and $j$ ($S_{ii}=0$). We then compute the symmetric normalized Laplacian matrix $A=I-D^{-\frac{1}{2}}SD^{-\frac{1}{2}}$, where  $D$ is a diagonal matrix, $D_{ii}=\sum_{j=1}^n {S_{ij}}$.

We train the network in a more limited manner, in a single phase without Fourier analysis, on 256000 problems. To avoid inverting singular matrices when training, we modify the Laplacian matrices to be non-singular by adding random positive values to the diagonal of the matrix, from distribution $U(0,0.2)$. Evaluation is done on the original singular matrices. To evaluate, we measure the number of LOBPCG iterations required to reach residual tolerance of $10^{-12}$ on a variety of problems, where the linear solver is a single W cycle. Table \ref{tab:spectral_clustering_iters} shows results on several distributions. Evidently, the network is able to generalize to different number of points, number of clusters, dimensions, and distributions.

\begin{table}[h]\scriptsize
    \centering
    \caption{Comparison of number of LOBPCG iterations required to reach specified tolerance in spectral clustering problems, averaged over 100 runs for each distribution}
    \vskip 0.15in
    \begin{tabular}{l|c|c|c|c}
        \toprule
        distribution & size & CAMG & ours & ratio\\
        \midrule
        two Gaussians          & $10^3$ & 15.67 & 13.44 & 85.8\%  \\
        two Gaussians          & $10^4$ & 20.95 & 18.82 & 89.8\%  \\
        two Gaussian 5-NN      & $10^3$ & 22.53 & 23.45 & 104.1\% \\
        five Gaussians         & $10^3$ & 19.99 & 17.41 & 87.1\%  \\
        two Gaussians 3D       & $10^3$ & 12.58 & 11.26 & 89.5\%  \\
        two moons              & $10^3$ & 23.44 & 21.47 & 91.6\%  \\
        two moons              & $10^4$ & 37.17 & 35.02 & 94.2\%  \\
        two concentric circles & $10^3$ & 19.48 & 16.84 & 86.5\%  \\
        \bottomrule
    \end{tabular}
    \label{tab:spectral_clustering_iters}
\end{table}

\paragraph{Ablation Study.}
We run a number of experiments to determine how the performance of the network is influenced by our design decisions. We evaluate performance on graph Laplacian problems, with networks trained with the following modifications: less message-passing layers, lower depth of MLPs, no concatenation of encoder features as input to message-passing layers, and no one-hot indicators on edge and node input features. Table \ref{tab:ablation} shows the success rate of these networks on problems with lognormal distribution of size 65536. As can be seen, performance moderately drops when lowering the depth of the MLPs or removing encoder concatenation, and significantly drops when lowering the number of message-passing layers, and removing indicator features.

\begin{table}[]\scriptsize
    \centering
    \caption{Success rate measured for graph Laplacian problems with lognormal distribution of size 65536. Tested on W-cycle, averaged over 100 runs for each architecture}
    \vskip 0.15in
    \begin{tabular}{l|c}
        \toprule
        architecture & success rate \\
        \midrule
        Suggested architecture   &  79\% \\
        Depth 2 MLP              &  74\% \\
        2 message-passing layers &  63\% \\
        No encoder concatenation &  75\% \\
        No indicator features    &  68\% \\
        \bottomrule
    \end{tabular}
    \label{tab:ablation}
\end{table}

\section*{Conclusion}
In this paper we propose  a framework for learning Algebraic  Multigrid (AMG) prolongation operators for linear systems which are defined directly on graphs, rather than on structured grids. We treat linear systems that can be expressed by sparse symmetric positive (semi-) definite matrices. We formulate the problem as a learning task and train a single graph neural network, with an efficient message-passing architecture, to learn a mapping from an entire class of such matrices to prolongation operators.  We employ an efficient and unsupervised  training on a limited class of block-circulant matrices. Our experiments indicate success, i.e, improved convergence rates compared to  classical AMG, on a variety of problems. This includes graph Laplacian problems over a triangulated mesh, where the edge weights are drawn randomly from some distribution,  diffusion partial differential equations discretized on 2D triangular meshes and spectral clustering problems. An interesting and important direction for future research is learning to select the coarse representatives as well as  the sparsity pattern of the prolongation matrix. 

 \section*{Acknowledgment}
This research was supported by the Israel Science Foundation, grant No. 1639/19.